\documentclass{article} % For LaTeX2e
\usepackage{nips14submit_e,times}
\usepackage{hyperref}
\usepackage{url}
\usepackage[pdftex]{graphicx}

\title{Real World Applications of Machine Learning Techniques over Large Mobile Subscriber Datasets}

\author{
Jobin Wilson, Chitharanj Kachappilly, Rakesh Mohan, Prateek Kapadia, Arun Soman\\
R\&D Department\\
Flytxt\\
Trivandrum-695581, India \\
\texttt\{jobin.wilson, chitharanj.kachappilly ,rakesh.mohan ,prateek.kapadia, arun.soman\}
\\@flytxt.com \\
\AND
Santanu Chaudhury \\
Department of Electrical Engineering \\
Indian Institute of Technology, Delhi \\
Hauz Khas, New Delhi-110 016, India \\
\texttt{santanuc@ee.iitd.ac.in} \\
}

\nipsfinalcopy % Uncomment for camera-ready version

\begin{document}

\maketitle
\begin{abstract}
Communication Service Providers (CSPs) are in a unique position to utilize their vast transactional data assets generated from interactions 
of subscribers with network elements as well as with other subscribers. CSPs could leverage its data assets for a gamut of applications such as 
service personalization, predictive offer management, loyalty management, revenue forecasting, network capacity planning, product bundle 
optimization and churn management to gain significant competitive advantage. However, due to the sheer data volume, variety, velocity and veracity
of mobile subscriber datasets, sophisticated data analytics techniques and frameworks are necessary to derive actionable insights in a 
useable timeframe. In this paper, we describe our journey from a relational database management system (RDBMS) based campaign management solution 
which allowed data scientists and marketers to use hand-written rules for service personalization and targeted promotions to a distributed 
Big Data Analytics platform, capable of  performing large scale machine learning and data mining to deliver real time service personalization,
predictive modelling and product optimization. Our work involves a careful blend of technology, processes and best practices, which facilitate 
man-machine collaboration and continuous experimentation to derive measurable economic value from data. Our platform
has a reach of more than 500 million mobile subscribers worldwide, delivering over 1 billion personalized recommendations 
annually, processing a total data volume of 64 Petabytes, corresponding to 8.5 trillion events.
\end{abstract}
\section{Introduction}
Telecommunications industry has emerged as an appropriate domain for applying large-scale data mining and machine learning techniques for a 
myriad of applications such as churn propensity scoring[4][5], fraud detection [6], improving customer relationship management [8][7],
network planning [9] and customer segmentation [10], due to availability of large volumes of high quality subscriber data [3][2].
Mobile internet connectivity is increasingly becoming ubiquitous. In 2011 itself, more than 50\% of all local searches were done from mobile devices.
In 2013, the total mobile phone subscriptions across the world crossed 6.5 billion, which accounts to nearly 92\% of the world's population [1].
A Gartner study predicts that by 2015-16, there will be more mobile devices connected to the Internet than desktops. These trends indicate that the 
Internet of Things (IoT) and the vision of connecting everyday objects eventually to the Internet is becoming as a reality [15]. Convergence of IoT, 
Big Data, and device interoperability through open standards has immense potential to affect every aspect of human life, with possibilities to
make virtually anything intelligent. However, this presents a completely new set of challenges to the current big data ecosystem. As granularity
of data increase and countless data sources generate streaming data in a variety of structured and unstructured formats, data mining algorithms need 
to adapt to handle an explosive volume of data in a variety of formats at varying velocity and veracity to generate value. The challenge will no 
longer be the absence of enough data volumes, but would be devising efficient algorithms, which can filter meaningful data from a vast ocean 
of raw data streams. Data mining challenges faced by today's CSPs serve as a good reference model that we could leverage to understand some of 
these futuristic challenges.

The purpose of this paper is two-fold. First, to introduce our real-time scalable machine learning and personalization platform architecture
and thereby contribute to the evolving best practices and guidelines around building and operationalizing large-scale analytics platforms. Second, to 
highlight a few specific use cases which utilize scalable analytics, and our key learnings from productionizing them. To this end, we intend to 
sketch our evolution from a relational database management system (RDBMS) based mobile marketing solution with limited data analytics 
capabilities and manual rule-based targeting into a scalable big data analytics platform, enabling large scale machine learning applications.
While the general practice of applying data mining techniques in an academic setting assume existence of a clearly defined mining problem and 
a ready-to-consume dataset along with well-defined performance metrics, it may be unrealistic for real-world problems. In practice, objectives 
and performance metrics may need to be derived from imprecisely defined business goals. We hope that this work will provide practical insights to
practitioners for engineering large analytics platforms and expose the data mining research community to novel challenges in productionizing 
machine learning algorithms.

\section{Our Evolution}
\label{sec:our_evolution}
In early days, marketing campaigns were akin to carpet bombing with only macro-level segmentation at best. Cumbersome data extraction processes,
multiple handoffs between various stakeholders, long lead times for execution, lack of response and campaign efficacy measurement, all pointed 
towards transformation to a fully integrated conversational marketing approach with impact measurement as a natural step in evolution. 
Our mobile marketing platform, with a host of applications like push marketing, interactive marketing and inventory management, focused on 
enabling this transformation for a wide range of business units, including Usage and Revenues (U\&R), Customer Acquisition, Customer Care and Retail.
This was realized by providing applications for seamless GUI driven campaign execution, automation of underlying processes, full integration to 
upstream and downstream systems, and closing the loop by providing feedback on campaigns, which could be leveraged in new campaign designs. 

Practical applications reinforced the need for specialized technical roles, like data management, incident management and data quality assurance 
that would support the marketer in his endeavor. For instance, adapting to changing file formats and managing erratic file delivery to assure 
data quality required dedicated technical personnel. Also, achieving operational efficiency was a challenge which required marketers, analysts
and system integrators to work in tandem. As marketing strategy evolved, marketers realized the value of data sciences as part of campaign design
and impact analysis. Real-time response based on subscriber actions, network events and subscriber location also emerged as prominent use cases. 
Later, analytics evolved beyond simple statistics on individual subscriber's key performance indicators (KPIs) and manually defined rules for grouping and classification, 
to machine learned tags, clustering, propensity analysis models and recommenders. As a result, the need for performance, tunable latencies and 
scalability increased multi-fold.

Our platform had to adapt from being a transformational campaign execution tool to a real-time data analytics powered mobile marketing ecosystem, 
catering to varied use cases and actors. Originally a fully RDBMS based solution, all data processing was performed within the database. 
As a first candidate for optimization, the Extract-Transform-Load (ETL) framework was identified, where the practice was to ingest data into a 
database and process using stored procedures. This approach was inefficient while processing call data record (CDRs) which may contain
billions of rows and hundreds of columns, often requiring cross references across multiple input sources. By externalizing basic 
transformations and data aggregation from RDBMS into Hadoop, we achieved performance improvements as significant as 100 times speedup in ETL 
and campaign target generation processes. Even then, we faced scalability challenges for CSPs with more than 50 million subscribers. 
This prompted us to overhaul our platform and practices to bring in the desired level of scalability, performance and resiliency.

\section{Analytics Platform Architecture}
\label{sec:analytics_platform_architecture}

For real-time delivery of machine learning outcomes at scale, our purpose-built architecture essentially decouples multiple layers 
and application flows. Computational layer, responsible for building and maintaining up to date models operate in a batch mode. Output from models
are persisted into a low latency key-value store at scheduled intervals, as configured. A service layer API provides 
real-time access to the generated insights. Computational layer is pluggable and allows co-existence of multiple underlying data processing
frameworks such as Apache Hadoop and Apache Spark, allowing model developers to make choices based on computational needs. Models are expressed as
workflows, using a domain specific language (DSL) based on XML, facilitating quick experimentation. Common feature engineering steps such as 
dimensionality reduction and sampling are built into the platform, as reusable components. Delivery channels make use of the 
personalization API to personalize offers and content across touchpoints. Our platform also
leverage RDBMS as transactional and metadata store, and an in-memory database for real time analytics. A host of specialized services are 
built in to the platform, as illustrated in the figure and summarized below.The platform allows other pluggable applications to consume these 
services, allowing a wide variety of business units to derive economic value from a common data pool.

\begin{itemize}
 \item Deployment and integrations - Built-in configurable interfaces to CSP's network elements which include various data sources, 
 communication channels, operations support and business support systems(OSS/BSS) and subscriber touch points; tools that speed up 
 deployments and ensure adherence to the reference architecture and best practices.
 
 \item Campaign management - Using a configurable rule management UI where domain experts could configure, 
 edit and manage huge number of rules against subscriber KPIs and insights 
 
 \item Analytics support - Using built-in/plug and play domain specific algorithms, data management services and adaptors,
 and a workflow orchestration mechanism
 
 \item Operations support - Using standardized logging, monitoring and alerting tools, and standard operating procedures. 
 We are also working on an anomaly detection framework that employs predictive analytics to enable proactive issue resolution.

\end{itemize}

\begin{figure}[h]
\label{fig:platform_architecture}
\begin{center}
      %\framebox[4.0in]{$\;$}
\includegraphics[width=4.5in]{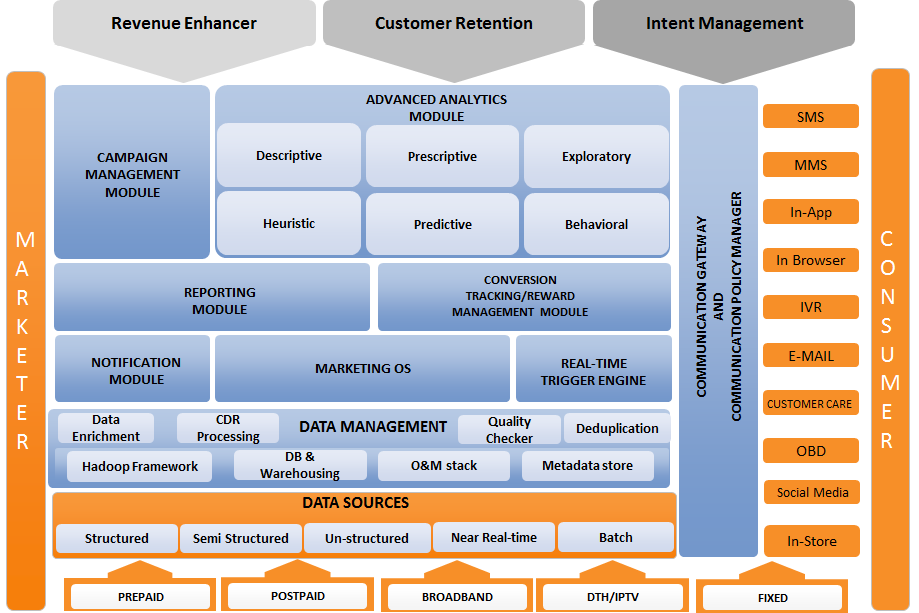}
\end{center}
\caption{Flytxt Platform Architecture}
\end{figure}

Lot of our design choices and component evolution were influenced by our learnings from our production systems.
For instance, while standard tools, such as Flume and Scoop, were available for data acquisition and ingestion into Hadoop, 
further data processing towards insight generation needed a custom data management and workflow orchestration framework. Producer-consumer 
interdependencies between workflows had to be modelled as data pipelines and processes that are triggered on satisfying a predicate 
(a combination of arrival of the required data, a schedule and/or the occurrence of an external event). The service also maintains a state 
against each process, file and insight at multiple check points, to provide visibility into their lifecycle status. Basic data quality checks 
like file consistency and header validations were also handled on receiving the file, done only once before supplying them to processes, 
thus avoiding repetition at the process level. Row level checks are usually disabled to avoid latencies, however data quality assurance
tools like daily trend reports, and trip wires based on accumulators and counters are provided to trigger alerts if deviations are observed 
beyond a threshold.  

We also observed that data access patterns varied across different processes, with different sources and sink. Some processes 
required low latency key-value lookups, as in the case of real-time trigger based rule evaluations, while others required accessing 
subsets of data qualifying a predicate. Some processes use iterative computing techniques which are I/O intensive, in worst case 
requiring multiple full table read-writes as in the case machine learning algorithms such as k-means. Applications such as targeting 
require a full table scan on the insight repository per target calculation, and write the result to an RDBMS.  Full table operations on 
the insight repository proved inefficient when multiple such application run together on a very large data set, as is the case with data 
written to RDBMS. A data access layer was thus introduced, which facilitates a publish-subscribe pattern, propagating data to all 
applications at varying latencies depending on the context. The layer also provides pluggable adaptors for data format conversions 
required for integrating external libraries, like Mahout, into the platform. For iterative computing, where 
repetitive disk operations were found to be wasteful, we introduced Apache Spark, which loads data in to a cluster's memory and allow 
repeated processing in-memory.
\section{Machine Learning Applications at Scale}
\label{sec:live_impact_stories}
In this section, we present a few interesting mobile marketing applications which make use of large scale machine learning techniques, 
which we have productionized successfully.
\subsection{Constrained fair ad-allocation for Mobile Advertising}
\label{subs:ad_allocation}
Matching potentially millions of subscribers to relevant mobile advertisements subject to constraints on advertiser budgets,
subscriber messaging limits, optimal revenue for the advertising network and fairness to all advertisers, along with the sheer data volumes 
and changing subscriber behavior, poses significant data management challenges. A direct, scalable solution to this problem could be extremely 
I/O intensive and suboptimal in infrastructure utilization, if not intractable. We make use of a near-linearly scalable approach to solving this 
constrained allocation problem using a combination of a scaling transformation which groups subscribers based on advertiser's target criteria and 
application of an integer programming technique.

Concretely, The ad-allocation problem can be summarized as follows. There exist a set of advertisers, with certain targeting criteria
expressed in terms of subscriber KPIs. These advertisers pay certain per unit price for every ad sent; they also have an allocation budget which 
denote the limit to the number of ads it can afford to send. These ads are to be allocated to a given set of subscribers, each having KPI values 
associated with them. Each subscriber also has a limit to the number of ads it is willing to receive, indicated by subscriber limit or frequency cap.
Ad-allocation is formulated as an optimization problem which could be solved using an integer program.

\begin{center}
$max$:  $\sum_j ~\left( p_j \sum_{i,e_{ij}=1} x_{ij}\right)$                          
\end{center}
~~~~~~~~~~~~~~~~~~Subject to:
\begin{center}
$\forall i : ~\sum_{j,e_{ij}=1} x_{ij} \leq  fcs_i$, frequency cap of $subscriber_i$\\
$\forall j : ~\sum_{i,e_{ij}=1} x_{ij} \leq  fcc_j$, frequency cap of $advertiser_j$\\
$\forall i,j : x_{ij} \in \{0,1\}$\\
\end{center}

here $\forall j, p_j$ denotes price paid by $advertiser_j$, per subscriber assigned to $advertiser_j$ and 
$e_{ij}$ indicate eligibility of $subscriber_i$ to be allocated to $advertiser_j$, determined by match between subscriber attribute values
and advertiser's target criteria; $x_{ij}$ indicates if $subscriber_i$ is allocated to $advertiser_j$. 

As we are dealing with millions of subscribers and thousands of advertisers, an efficient representation of the problem is critical.
Our approach involved grouping subscribers based on advertiser's target criteria so that all allocations could be made in such a way that a certain 
volume of these subscriber groups would be allocated to each advertiser, thus converting a large optimization problem to a smaller problem 
with fewer variables.

A subscriber group, $G$, is characterized by the following properties.
\begin{itemize}
 \item $G$ is a subset of the entire subscriber base
 \item For every other group $G'$, $G \cap G' = \emptyset $ , That is, all the groups are mutually exclusive.
 \item All the subscribers in $G$, are targeted by exact same set of advertisers.
 \item The difference between frequency caps which represent individual limit on the number of ads receivable per day, of any 2 subscribers in $G$ 
 is at most 1.
 \end{itemize}

In addition to the above aspects, every subscriber must belong to at least one group. The group allocation problem corresponding to a 
subscriber-advertiser allocation problem is obtained by reading the advertisers requirements. The attribute requirements specified by all 
the advertisers are gathered first. Every eligibility criterion is represented by a bit. For all the subscribers satisfying a criterion, the 
corresponding bit is set to one. Subscribers are further grouped based on their frequency caps as above. This way, we achieve a set of subscriber
groups satisfying all the above properties. All the members of a group are targeted by the exact same set of advertiser, because of the bit
representation. We define frequency cap of a subscriber group $G$ to be equal to the sum of frequency caps of all the subscribers in that group.
These formulated groups, along with the defined frequency cap properties are used to construct a group allocation problem, 
and solved for a feasible solution. This defines the procedure to derive a group allocation problem from an instance of the subscriber-advertiser
allocation problem. In this approach, subscriber attributes are read only once from the data store and we explicitly need not care about frequency 
cap feasibility as it would be taken care of by the model itself. Once a feasible solution to the group allocation problem is obtained,
we derive the actual allocation of individual subscribers to advertisers by iterating over each group. Within each group, we arrange subscribers 
in the descending order of frequency cap and allocation to each advertiser is carried out in such a way that subscribers with higher frequency 
cap gets allocated first. Fairness is an important aspect in ad-allocation. We require that no advertiser must starve because it pays less.
Fairness is easily introduced into the model through additional constraints to the optimization problem so that ad allocations are made to
advertisers proportional to the price they are willing to pay per impression.

We make use of a mixed integer linear program solver library called SYMPHONY[11] to solve the formulated problem.
A comparative study of solution running time and problem size of the original subscriber-advertiser allocation problem to
the group allocation problem, is described in Table \ref{scaledVsUnscaled}, demonstrating the effect of our reduction approach. 
The problem size indicated here is the size of the formulated .mps file representing the integer program that we provide to the solver.

\begin{table}[htp]      
\caption{Comparison Of Original To The Reduced Allocation}
\centering
\begin{tabular}{c c c c c}
\hline\hline
Number of &  Average & & Integer & Time to\\
advertisers/ & number of & \raisebox{1.0ex}{Problem} & Program(IP) & solve\\
subscribers & Targets & \raisebox{1.0ex}{Type} & size & the IP\\[0.5ex]
\hline\hline
\\[0.5ex]
152/& & Unscaled & 844MB & 813s\\[-1ex]
16 million & \raisebox{1.5ex}{100,000}&Scaled
& 188kB & 6.2ms\\[1ex]
\hline\\[1.0ex]
35/& & Unscaled & 42GB & unsolved\\[-1ex]
12 million & \raisebox{1.5ex}{6,000,000}&Scaled
& 84kB & 1.6ms
\\[1ex]
\hline\hline
\end{tabular}
\label{scaledVsUnscaled}
\end{table} 

While operationalizing this model, we realized that our reduction approach gave us significant gains. As solving large scale optimization
problems are computationally challenging, reducing it to a smaller problem without compromising the quality of solution is an ideal approach 
in this context. Subscriber KPI aggregations were performed in a distributed manner using map-reduce jobs, however the core ad-allocation algorithm
executes as a non-distributed process, once the scaling transform is performed. As ad delivery is a continuous automated process, we built a 
workflow which made use of the platform components described in \ref{sec:analytics_platform_architecture}, and scheduled it to precompute and persist
ranked advertisements daily, for each subscriber.

\subsection{Content Personalization using Topic Modelling}
\label{subs:portal_personalization}
CSPs offer a lot of multimedia content with varying attributes, across its touchpoints. Constructing a concise and interpretable subscriber 
profile from a subsriber's purchase history across touchpoints was challenging. We implemented a scalable hybrid model for content personalization 
by combining standard content based filtering algorithm and latent Dirichlet allocation (LDA) [17]. Content providers generally add meta data 
in text form and tags to the content. Additionally, public data sources such as Wikipedia may be used to enrich meta data on popular content. Our 
approach involves transforming subscribers and contents into a single latent topic space to generate recommendations.

Each content is represented as a text document containing meta data about the content, to form a corpus of documents. LDA 
is performed on this corpus to discover document-topic probability distribution as well as topic-word probability distribution. Document-topic 
distribution serves as a content profile, where each topic probability is a feature, indicating how strongly that topic describes the content.
We sum up document-topic distributions scaled by normalized subscriber rating, to generate subscriber topic-distribution vector in the same 
latent topic space, which serves as a consistent subscriber profile across touchpoints. Though topics are latent, most frequent keywords 
corresponding to each topic allows us to interpret the content profiles and subscriber profiles constructed by this model.

Once all subscriber profiles and content profiles are expressed in a common feature space, similarity of a subscriber to another subscriber or to
a content can be easily calculated using symmetric Kullback-Leibler divergence between their corresponding latent topic distributions.
%\begin{center} 
 $S_{topic}(U,U')$=$e^{-D_{\mathrm{KLSymmetric}}(U_f,U'_f)}$ \\\\
 $D_{\mathrm{KLSymmetric}}(U_f,U'_f)= D_{\mathrm{KL}}(U_f || U'_f)+ D_{\mathrm{KL}}(U'_f || U_f)$ \\ \\
 $D_{\mathrm{KL}}(U_f || U'_f) = \sum_i \ln\left( \frac{U_f(i)}{U'_f(i)}\right)U_f(i)$ \\\\
%\end{center}
Here $U_f$ denotes topic distribution representing subscriber $U$'s discovered  profile.
Symmetric Kullback-Leibler divergence between latent topic distributions is converted into a similarity score 
using an exponential function as indicated above, to ensure that the similarity value lies within the interval [0,1]. 
Previous work of Wilson et al. describes this approach in detail [16].

Our initial approach involved constructing a non-distributed user-neighborhood based recommender which used this custom user-similarity function
as opposed to the rating overlap based similarity calculation which is common in standard user-based Collaborative Filtering (CF) algorithms. In our approach,
recommendations for each subscriber is generated by first forming a candidate list of all distinct contents that at least one subscriber in the 
current subscriber's neighborhood has accessed. This list is then sorted based on the content popularity within the neighborhood, calculated as the
fraction of neighbors who preferred that content, and $K$ most popular contents are recommended. Our benchmark studies on Movielens 1M dataset [12] 
along with IMDB dataset from IMDB interfaces [13] indicate that this approach significantly outperforms standard implementations of user-based CF and 
item-based CF in Apache Mahout, in terms of classification accuracy metrics such as precision, recall and F-measure. Table \ref{tab:tab_dataset_properties} describe the 
properties of the dataset used for this study. F-measure analysis from our cross-validation experiments is plotted below.

\begin{table}
\renewcommand{\arraystretch}{1.3}
\caption{Dataset Properties}
\label{tab:tab_dataset_properties}
\centering
\begin{tabular}{c|c|c|c|c}
    \hline
    Dataset  &  Users  & Items  &  Max. Ratings Per User  &  Avg. Ratings Per User\\
    \hline
    \hline
   Movielens(Training)
 &  6040  &  3677  &  1851  &  132.48\\

    \hline
   Movielens(Testing) &	6040 &	3468 &	462 &	32.11\\
   \hline
\end{tabular}
\end{table}

\begin{figure}[h]
\label{fig_movielens_irstats}
\begin{center}
  %\framebox[4.0in]{$\;$}
\includegraphics[width=3.5in]{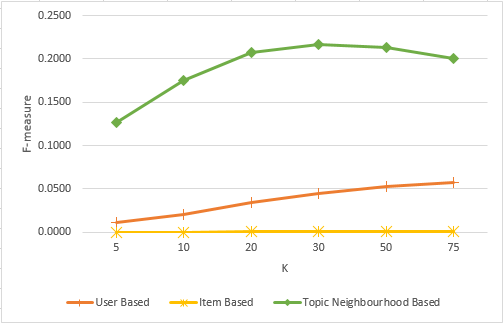}
\end{center}
\caption{F-measure Analysis: Movielens 1M}
\end{figure}
To scale up the implementation, we split the recommender system into three phases. First phase generates the content-topic probability distributions
or content profiles along with top keywords corresponding to each topic. We make use of CVB algorithm which is implemented in Apache Mahout for
LDA [14]. Second phase aggregates content-topic distributions according to subscriber rating data to generate subscriber-topic distributions or 
profiles, using a map-reduce job. Third phase generates the actual recommendations per subscriber based on the learned profiles using 
multiple map-reduce jobs. Final output from the model is generated as a flat file containing $K$ recommendations per subscriber, which gets 
persisted into a low-latency key-value store for real-time personalization delivery. Seperate workflows were created to automate this whole process.
First phase got scheduled to run weekly, as new contents got added weekly. Second and third phase were daily jobs, due to regular usage. 

Initially, content personalization was limited to a few touchpoints and hence subscriber base actively using it was limited. As the usage increased, 
calculation of user-similarity matrix became computationally intensive. Also, frequent profile updates quickly made the similarity matrix obsolete, 
forcing its frequent re-computation. To solve this problem, efforts are underway to modify the ranking logic of our recommender system to estimate 
similarity between subscriber profiles and content profiles directly using Kullback-Leibler Divergence.

During the course of operationalizing this model, we realized that a proper breakup of a recommender model into decoupled phases allows 
scaling them independently. Also, having a manual rule-based recommendation mechanism integrated with the recommender system is practically useful
as marketers could optionally override machine generated recommendations based on business priorities and specialized domain knowledge. Manual rules
proved to be also useful to generate default recommendations in situations where enough information about subscribers or content to be recommended
is unavailable. Another aspect that we realized was that parallelizing a process in itself may not be sufficient to meet scalability challenges
posed by real-world problems as in this case, where revisiting the ranking logic was an appropriate choice.

\section{Lessons Learned From Our Journey}
\label{sec:conclusion}
Operationalizing large-scale data mining platforms for real-world CSP applications are more than trivial and require a correct blend of technology,
human resource, organizational structures and organizational processes. We realized that, so far, there is no single Big Data technology,
which can readily cater to all CSP needs. 

Practically, a purpose built hybrid architecture, which allows co-existence of several big data and conventional data management technologies
proved to be effective. Analytics platform should support multiple data organization strategies to facilitate the necessary data access patterns.
Providing pluggable support for multiple distributed and non-distributed execution modalities such as map-reduce and in-memory processing gives 
flexibility to the analytics process to choose the appropriate execution framework depending on data volume and nature of computation. 
Need for a meaningful logging strategy coupled with a centralized system that derives operational insights, is paramount. Adopting open-source 
technologies and extending them as per needs, proved to be of immense value in terms of reducing feature delivery time and total cost of ownership.

Another set of key learnings were around data management. A proper data quality assurance framework is essential to ensure data sanity within 
the platform. We also recognized the importance of a data lifecycle management system, which tracks data flows across the platform. Maintaining 
a data catalogue which captures details of data sources, file formats, derived KPIs and insights was observed to be helpful for seamless 
integration and consumption of insights across the platform. Our scaling efforts on data mining algorithms taught us that there is no single 
scaling-recipe which is universal. Data preprocessing and feature engineering through a careful blend of domain knowledge and data sciences may 
prove to be a lot more important than the sophistication of learning algorithms, in practice.

\section{References}

\small{

[1] Portio Research Ltd. Portio Research Mobile Factbook 2013,
Report accessed online on July 21,2014 : http://www.portioresearch.com/en/free-mobile-factbook.aspx

[2] Carbonara, Leo, Huw Roberts, and Blaise Egan. "Data mining in the telecommunications industry." 
Lecture notes in computer science (1997): 396-396.

[3] Weiss, Gary M. "Data mining in telecommunications." 
Data Mining and Knowledge Discovery Handbook. Springer US, 2005. 1189-1201.

[4] Hung, Shin-Yuan, David C. Yen, and Hsiu-Yu Wang. "Applying data mining to telecom churn management." 
Expert Systems with Applications 31.3 (2006): 515-524.

[5] Wei, Chih-Ping, and I. Chiu. "Turning telecommunications call details to churn prediction: a data mining approach." 
Expert systems with applications 23.2 (2002): 103-112.

[6] Phua, Clifton, et al. "A comprehensive survey of data mining-based fraud detection research." 
arXiv preprint arXiv:1009.6119 (2010).

[7] Daskalaki, Sophia, et al. "Data mining for decision support on customer insolvency in telecommunications business."
European Journal of Operational Research 145.2 (2003): 239-255.

[8] Rygielski, Chris, Jyun-Cheng Wang, and David C. Yen. "Data mining techniques for customer relationship management." 
Technology in society 24.4 (2002): 483-502.

[9] Gawrysiak, Piotr, and Michal Okoniewski. "Applying data mining methods for cellular radio network planning." Intelligent Information Systems. 
Physica-Verlag HD, 2000. 87-98.

[10] Gilbert, A. Lee, and Jon D. Kendall. "A marketing model for mobile wireless services." System Sciences, 2003. Proceedings of the 36th Annual Hawaii 
International Conference on. IEEE, 2003.

[11] COIN-OR Symphony development home page. https://projects.coin-or.org/SYMPHONY, Last accessed on 22nd July 2013.

[12] Riedl, J., and J. Konstan. "Movielens dataset." (1998).

[13] IMDb., "Internet movie database." , February 2014. [Online]. Available: http://www.imdb.com/interfaces

[14] S. Owen, R. Anil, T. Dunning, and E. Friedman, Mahout in action. Manning, 2011.

[15] Chaouchi, Hakima. "Introduction to the Internet of Things." The Internet of Things: Connecting Objects to the Web (2010): 1-33.

[16] Wilson, Jobin, Santanu Chaudhury, and Brejesh Lall. "Improving Collaborative Filtering based Recommenders using Topic Modelling." 
Proceedings of the 2014 IEEE/WIC/ACM International Joint Conferences on Web Intelligence (WI) and Intelligent Agent Technologies (IAT)-Volume 01.IEEE Computer Society, 2014.

[17] D. M. Blei, A. Y. Ng, and M. I. Jordan, "Latent dirichlet allocation", the Journal of machine Learning research, vol. 3, pp. 993?1022, 2003.
}

\end{document}